\documentclass[9pt,arxiv]{lapreprint}

\usepackage[version=4]{mhchem} 
\usepackage{siunitx}    
\usepackage{pdflscape}  
\usepackage{rotating}   
\usepackage{textgreek}  
\usepackage{gensymb}    
\usepackage[misc]{ifsym} 
\usepackage{listings}   
\usepackage{colortbl}   
\usepackage{tabularx}   
\usepackage{longtable}  
\usepackage{subcaption}
\usepackage{multirow}
\usepackage{snotez}     
\usepackage{csquotes}   

\usepackage{amsmath}
\usepackage{mathtools}
\usepackage{enumitem}
\usepackage{cancel}
\usepackage{todonotes}

\usepackage{amsmath,amsfonts,bm}

\def\eqref#1{equation~\ref{#1}}
\def\Eqref#1{Equation~\ref{#1}}

\def\1{\bm{1}}

\def\rvepsilon{{\mathbf{\epsilon}}}

\def\vtheta{{\bm{\theta}}}

\def\vx{{\bm{x}}}

\def\vz{{\bm{z}}}

\DeclareMathAlphabet{\mathsfit}{\encodingdefault}{\sfdefault}{m}{sl}
\SetMathAlphabet{\mathsfit}{bold}{\encodingdefault}{\sfdefault}{bx}{n}

\newcommand{\E}{\mathbb{E}}

\newcommand{\KL}{D_{\mathrm{KL}}}

\DeclareMathOperator*{\argmin}{arg\,min}

\DeclarePairedDelimiterX{\infdivx}[2]{(}{)}{%
  #1\;\delimsize\|\;#2%
}
\renewcommand{\KL}{\mathrm{KL}\infdivx}

\newcommand{\ELBO}{\mathrm{ELBO}}

\newcommand{\fracpartial}[2]{\frac{\partial #1}{\partial  #2}}

\usepackage[			
	backend=biber,      
        style=authoryear,   
	natbib=true,		
	hyperref=true,	    
	alldates=year,      
]{biblatex}

\title{A Tutorial on Parametric\\Variational Inference}

\author[ 
        1 \Letter]{Jens Sjölund}

\affil[1]{Department of Information Technology, Uppsala University, Sweden}

\correspondence{jens.sjolund@it.uu.se}{}

\funding[]{This work was partially supported by the Wallenberg AI, Autonomous Systems and Software Program (WASP) funded by the Knut and Alice Wallenberg Foundation.}

\leadauthor{Sjölund}
\shorttitle{A Tutorial on Parametric Variational Inference}

\begin{document}
\maketitle

\begin{abstract}
  Variational inference uses optimization, rather than integration, to approximate the marginal likelihood, and thereby the posterior, in a Bayesian model. Thanks to advances in computational scalability made in the last decade, variational inference is now the preferred choice for many high-dimensional models and large datasets. This tutorial introduces variational inference from the parametric perspective that dominates these recent developments, in contrast to the mean-field perspective commonly found in other introductory texts.   
\end{abstract}

\section{Introduction}
\looseness=-1
In Bayesian machine learning and statistics, the central object of interest is the posterior distribution found by Bayesian inference---combining prior beliefs with observations according to Bayes' rule. In simple cases, such as in conjugate models, this can be done exactly. But, general (non-conjugate) models require approximate inference techniques such as Monte Carlo or variational inference. These have complementary strengths and weaknesses, hence the most appropriate choice is application dependent. We focus on variational inference, which is on the one hand \emph{not} guaranteed to be asymptotically exact but is on the other hand computationally efficient and scalable to high-dimensional models and large datasets. 

\paragraph{Notation}
We use a single observation variable $\vx$ to denote both the observed inputs and outputs. Our primary interest is however in the latent variables $\vz$. Since we adhere to the Bayesian framework, the ``parameters'' of a model (such as the slope and intercept in a linear regression model) that are assigned priors are actually latent variables. We denote remaining parameters of interest by $\vtheta$.

\section{Variational inference}
So, why do we need variational inference? First, recall that to infer anything about the latent variables from our observations, we need the posterior:
\begin{align}
    p_\vtheta(\vz \mid \vx) = \frac{p_\vtheta(\vx, \vz)}{p_\vtheta(\vx)}.
\end{align}
The expression in the denominator, $p_\vtheta(\vx)$, is called the marginal likelihood of $\vx$ because it can be rewritten as a marginalization over the latent variables:
\begin{align}
    p_\vtheta(\vx) = \int p_\vtheta(\vx, \vz)\,d\vz.\label{eq:marginal_likelihood}
\end{align}
The catch is that in practice this integral is often intractable, i.e. not computable in closed form. Since $\vx$ are our observations, the marginal likelihood is a (normalizing) constant. Nevertheless, without knowing this constant the utility of the posterior is limited. Hence the need for approximate inference.

\paragraph{Key idea}
The key idea in variational inference is to replace the intractable marginal likelihood with a tractable lower bound that we then maximize. Modeling mainly consists of choosing a family $\mathcal{Q}$ of probability distribution that are well-behaved yet sufficiently expressive. More specifically, we want there to be a distribution $q\in \mathcal{Q}$, called the variational posterior, that can be used as a drop-in replacement for the true posterior. The variational posterior should therefore be ``close'' to the true posterior $p_\vtheta(\vz\mid \vx)$ and at the same time (relatively) easy to find. The search procedure amounts to mathematical optimization, which is why variational inference is sometimes described as trading a difficult integration problem for an easier optimization problem. 

\paragraph{The evidence lower bound (ELBO)}
In variational inference, the distance between the true posterior $p(\vz\mid \vx)$ and the variational posterior $q(\vz)$ is measured using the Kullback-Leibler (KL) divergence,
\begin{equation}
    \KL{q(\vz)}{p(\vz\mid \vx)}
    =-\int q(\vz)\log \left(\frac{p(\vz\mid \vx)}{q(\vz)}\right) d\vz.\label{eq:KL}
\end{equation}
Other distance measures can also be used to make the variational posterior similar to the true posterior, but the KL divergence has a particular benefit: through a neat trick we can simultaneously estimate the marginal likelihood and circumvent the need to evaluate the posterior in \eqref{eq:KL}. To see how, we first note the two mathematical identities:
\begin{align}
    &\int q(\vz) \,d\vz = 1,\label{eq:unity}\\
    &p(\vx)=\frac{p(\vx,\vz)}{p(\vz\mid\vx)}
    =\frac{p(\vx,\vz)}{q(\vz)}\left(\frac{p(\vz\mid\vx)}{q(\vz)}\right)^{-1}.\label{eq:rearrange}
\end{align}
Using these we may rewrite the marginal likelihood as follows:
\begin{align}
    \log p(\vx) &\underset{(\ref{eq:unity})}{=} \log p(\vx) \cdot \int q(\vz) \,d\vz= \int q(\vz) \log p(\vx) \,d\vz\nonumber\\
    &\underset{(\ref{eq:rearrange})}{=}\int q(\vz) \log \left(\frac{p(\vx,\vz)}{q(\vz)}\right) d\vz- \int q(\vz) \log \left(\frac{p(\vz\mid\vx)}{q(\vz)}\right)d\vz\nonumber\\
    &=\int q(\vz) \log \left(\frac{p(\vx,\vz)}{q(\vz)}\right) d\vz + \KL{q(\vz)}{p(\vz\mid\vx)}.\label{eq:evidence_decomposition_long}
\end{align}
Because the KL divergence is always nonnegative, the first term lower bounds the log marginal likelihood (also known as the evidence) for any $q$, and is therefore known as the evidence lower bound (ELBO):
\begin{equation}
    \ELBO(q(\vz))=\int q(\vz) \log \left(\frac{p(\vx,\vz)}{q(\vz)}\right) d\vz = \E_{q(\vz)}\left[\log p(\vx,\vz) - \log q(\vz)\right].\label{eq:elbo}
\end{equation}
\Eqref{eq:evidence_decomposition_long} can thus be written more succinctly as
\begin{equation}
    \log p(\vx) = \ELBO(q(\vz))+\KL{q(\vz)}{p(\vz\mid\vx)}.\label{eq:evidence_decomposition}
\end{equation}
For a fixed model $p(\vx,\vz)$, the (log) evidence is a constant. Hence---recalling that the KL divergence is nonnegative---we conclude that maximizing the ELBO is equivalent to minimizing the KL divergence. This is great, because to compute the KL divergence we would have to marginalize over a function that includes the same intractable posterior that we want to estimate. In contrast, the model only enters in the ELBO through the joint distribution $p(\vx,\vz)$, which means that, first, we don't need to compute the problematic integral in \eqref{eq:marginal_likelihood} and, second, we can factorize the joint distribution, e.g., as encoded by a directed graphical model \citep{wainwright2008graphical}.

\begin{featurebox}
\caption{}\label{box:gamma_lognormal}
Suppose we have a single observation $x$ from an Exp($\lambda$) likelihood with a Gamma($\alpha$, $\beta$) prior on the rate parameter $\lambda$. Assuming that $\alpha$ and $\beta$ are known, the only latent variable of interest is $z=\{\lambda\}$. Specifically,
\begin{align*}
    p(x \mid \lambda) &= \lambda e^{-\lambda x},\\
    p(\lambda) &= \frac{\beta^\alpha}{\Gamma(\alpha)}\lambda^{\alpha-1} e^{-\beta \lambda},
\end{align*}
where $\Gamma(\alpha)$ is the Gamma function. Since the Gamma distribution is the conjugate prior for $\lambda$, we know that the posterior is also a Gamma distribution. Invoking Bayes' rule and disregarding all factors not including $\lambda$, we find that
\begin{equation*}
    p(\lambda\mid x) \propto p(x \mid \lambda) p(\lambda) \propto \lambda^\alpha e^{-\lambda(\beta+x)}.
\end{equation*}
Hence, we identify the posterior as $p(\lambda\mid x)=\text{Gamma}(\alpha+1, \beta+x)$.

But, let's pretend we don't know this and instead want to fit a Lognormal($\mu$, $\sigma^2$) distribution to the posterior using variational inference, i.e.
\begin{equation*}
    q(\lambda)=\frac{1}{\lambda\sigma\sqrt{2\pi}}\exp\left(-\frac{(\log \lambda - \mu)^2}{2\sigma^2}\right).
\end{equation*}
From \eqref{eq:elbo} we have that
\begin{align*}
    &\ELBO(q(\lambda)) = \E_{q(\lambda)}\left[\log \left(p(x\mid\lambda)p(\lambda)\right) - \log q(\lambda)\right]\\
    &=\E_{q(\lambda)}\left[\log \left(\frac{\beta^\alpha}{\Gamma(\alpha)}\right) + \alpha\log\lambda - \lambda(\beta+x) + \log \sqrt{2\pi} + \log \sigma + \log \lambda + \frac{(\log \lambda-\mu)^2}{2\sigma^2}\right]\\
    &= \log\left(\frac{\beta^\alpha\sqrt{2\pi}}{\Gamma(\alpha)}\right) + (\alpha+1) \E_{q(\lambda)}\left[\log \lambda\right]-(\beta+x)\E_{q(\lambda)}\left[\lambda\right]+\log \sigma+\frac{1}{2\sigma^2}\E_{q(\lambda)}\left[(\log \lambda-\mu)^2\right].
    %
\end{align*}
The expectation $\E_{q(\lambda)}\left[\lambda\right]=\exp\left(\mu+\frac{\sigma^2}{2}\right)$ since, by definition, it is the mean of the lognormal distribution. Furthermore, the change of integration variables $y=\log\lambda$, which transforms  $q(\lambda)$ into $q(y)=\mathcal{N}(\mu,\sigma^2)$, shows that
\begin{align*}
    \E_{q(\lambda)}\left[\log \lambda\right]& = \E_{q(y)}\left[y\right]=\mu,\\
    \E_{q(\lambda)}\left[(\log \lambda-\mu)^2\right] &= \E_{q(y)}\left[(y-\mu)^2\right]=\sigma^2.
\end{align*}
The final expression for the ELBO is thus
\begin{equation*}
    \ELBO(q(\lambda))= \log\left(\frac{\beta^\alpha\sqrt{2\pi}}{\Gamma(\alpha)}\right) + (\alpha+1) \mu -(\beta+x)e^{\mu+\frac{\sigma^2}{2}}+\log \sigma+\frac{1}{2}.
\end{equation*}

\end{featurebox}

\paragraph{Modeling}
How, then, do we choose the variational family $\mathcal{Q}$? Historically, the dominant approach has been to assume a particular \emph{factorization} of the variational posterior, and to use calculus of variations to search for distributions that match this factorization. This is known as  mean-field variational inference \citep{blei2017variational}, and is still the approach most-often taught in classes. However, mean-field variational inference is only applicable to a rather limited set of models. Most of the successes of VI in the last 10--15 years have instead taken a parametric approach, where the variational family is \emph{parameterized} by a highly expressive model such as a deep neural network. One can then use ``standard'' optimization techniques to search for the parameters $\vtheta^*$ that maximize the ELBO.  In light of the above, this tutorial focuses exclusively on parametric variational inference.

In example \ref{box:gamma_lognormal}, we indeed took the parametric approach, since the variational posterior was explicitly parameterized by a Lognormal distribution with parameters $\vtheta=\{\mu,\sigma\}$. In example \ref{box:elbo}, we take a closer look at the ELBO for a specific instance of this model. 

To approximate the true posterior distribution accurately, we want the variational family $\mathcal{Q}$ to be as rich as possible so long as we maintain tractability---it is impossible to overfit! However, as example \ref{box:mismatched_support} shows, there is one pitfall to be aware of: \emph{$q(\vz)$ needs to be zero whenever $p(\vz\mid\vx)$ is zero}.

\begin{featurebox}
\caption{}\label{box:elbo}
To make thing more concrete, we continue with the setting from example \ref{box:gamma_lognormal} and set $\alpha=3$, $\beta=1$, and $x=1$.
The evidence $p(x)$ is the, previously neglected, proportionality constant relating the posterior and the joint distributions,
\begin{equation*}
    p(x)=\frac{p(x,\lambda)}{p(\lambda\mid x)}=\frac{\Gamma(\alpha+1)}{\Gamma(\alpha)}\frac{\beta^\alpha}{(\beta+x)^{\alpha+1}}=\frac{\alpha\beta^\alpha}{(\beta+x)^{\alpha+1}}.
\end{equation*}
Inserting the numerical values above gives $p(x=1)=3/16$. 

For simplicity, we fix $\sigma=0.5$ in the variational posterior (this corresponds approximately to the value found by moment matching) and study the effect of changing $\mu$.

\begin{center}
    \includegraphics[width=0.65\textwidth]{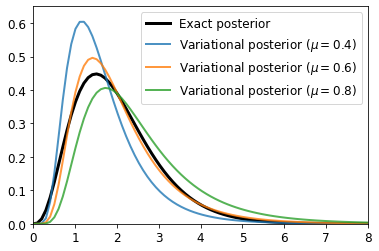}
\end{center}
\featurefig{The fit of a Lognormal($\mu$, $\sigma^2=0.25$) variational posterior to a Gamma($4$, $2$) posterior for different values of $\mu$.} 

\begin{center}
    \includegraphics[width=0.65\textwidth]{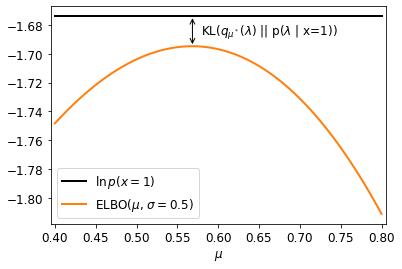}
\end{center}
\featurefig{How well the ELBO approximates the log evidence depends on the parameter $\mu$. The gap corresponds exactly to the KL divergence, hence maximizing the ELBO is equivalent to minimizing the KL divergence.}

\end{featurebox}

\begin{featurebox}
\caption{}\label{box:mismatched_support}
Let's return to Example \ref{box:gamma_lognormal} and see what happens if we try to use an $\mathcal{N}(\mu, \sigma^2)$ distribution as the variational posterior, i.e.
\begin{equation*}
    q(\lambda) = \frac{1}{\sqrt{2\pi\sigma^2}}\exp\left(-\frac{(\lambda-\mu)^2}{2\sigma^2}\right).
\end{equation*}
Deriving the ELBO as before, we have that
\begin{align*}
    &\ELBO(q(\lambda)) = \E_{q(\lambda)}\left[\log \left(p(x\mid\lambda)p(\lambda)\right) - \log q(\lambda)\right]\\
    &=\E_{q(\lambda)}\left[\log \left(\frac{\beta^\alpha}{\Gamma(\alpha)}\right) + \alpha\log\lambda - \lambda(\beta+x) + \log \sqrt{2\pi} + \log \sigma + \frac{(\lambda-\mu)^2}{2\sigma^2}\right]\\
    &= \log\left(\frac{\beta^\alpha\sqrt{2\pi}}{\Gamma(\alpha)}\right) + \alpha \underbrace{\E_{q(\lambda)}\left[\log \lambda\right]}_{\text{undefined!}}-(\beta+x)\underbrace{\E_{q(\lambda)}\left[\lambda\right]}_{=\mu}+\log \sigma+\frac{1}{2\sigma^2}\underbrace{\E_{q(\lambda)}\left[(\lambda-\mu)^2\right]}_{=\sigma^2}
\end{align*}
The logarithm is only defined for positive values, hence $\E_{q(\lambda)}\left[\log \lambda\right]$ is undefined. This illustrates an important caveat when choosing the variational distribution: \emph{$q(\vz)$ needs to be zero whenever $p(\vz\mid\vx)$ is zero}.
\end{featurebox}

\paragraph{Estimating the ELBO}
\looseness=-1
In the examples we've seen so far the expectations could be computed in closed form. But that will rarely be the case in general (non-conjugate) models. We can, however, use a Monte Carlo estimate to replace the expectation with a sum,
\begin{align}
     \ELBO(q(\vz))&=\int q(\vz) \log \left(\frac{p(\vx,\vz)}{q(\vz)}\right) d\vz = \E_{q(\vz)}\left[\log p(\vx,\vz) - \log q(\vz)\right]\nonumber\\
     &\approx \frac{1}{L}\sum_{i=1}^L \left(\log p(\vx,\vz^{(i)}) - \log q(\vz^{(i)})\right).
\end{align}
The key requirement is that we are able to draw samples $\vz^{(i)}$ from the variational posterior $q(\vz)$. But, as suggested by the previous section, it is not enough to evaluate the ELBO for a given $q\in\mathcal{Q}$---we want to find the best $q$! Having parameterized the variational posterior $q_\vtheta(\vz)$ with the parameters $\vtheta$, we may rephrase this as finding parameter values that maximize the ELBO. For efficient optimization, however, we need to evaluate both the objective function (the ELBO) \emph{and its gradient}.

\section{Gradient-based optimization of the ELBO}
In optimization, it is standard practice to consider minimization problems. (Since a maximization problem can be transformed into a minimization problem by negating the objective function, this can be done without loss of generality.) We thus express our optimization problem as:
\begin{align}
    \vtheta^* 
    &= \underset{\vtheta}{\argmin} -\E_{q_\vtheta(\vz)}\left[\log p(\vx,\vz) - \log q_\vtheta(\vz)\right].\label{eq:elbo_opt}
\end{align}
Applying, for instance, gradient descent to this problem corresponds to the iterations
\begin{equation}
    \vtheta_{k+1} = \vtheta_k + \eta \nabla_\vtheta \E_{q_\vtheta(\vz)}\left[\log p(\vx,\vz) - \log q_\vtheta(\vz)\right], \qquad k=0, 1, \ldots
\end{equation}
where the hyperparameter $\eta > 0$ is the step size. But this reveals a complication: the gradient acts on the parameters of the distribution that we compute the expectation over. Consequently, we cannot simply move the gradient inside the expectation, nor can we use the Monte Carlo trick to first replace the expectation with samples and then compute the gradient on those. But there are other, less direct, ways of applying the Monte Carlo idea that do work (incidentally, this turns gradient descent into stochastic gradient descent). We begin by rewriting the gradient of the ELBO as follows \citep{ranganath2014black}:
\begin{align}
   &\nabla_\vtheta\E_{q_\vtheta(\vz)}\left[\log p(\vx,\vz) - \log q_\vtheta(\vz)\right] =
   \nabla_\vtheta \int\left(\log p(\vx,\vz) - \log q_\vtheta(\vz)\right) q_\vtheta(\vz)\,d\vz\nonumber\\
   &=\int \left(\log p(\vx,\vz) - \log q_\vtheta(\vz)\right)\nabla_\vtheta \, q_\vtheta(\vz)\,d\vz-\int \left(\nabla_\vtheta \log q_\vtheta(\vz)\right) q_\vtheta(\vz)\,d\vz
\end{align}
But the second term in this expression vanishes,
\begin{align}
    &\int \left(\nabla_\vtheta \log q_\vtheta(\vz)\right)q_\vtheta(\vz)\,d\vz = \int \frac{\nabla_\theta q_\vtheta(\vz)}{\cancel{q_\vtheta(\vz)}} \cancel{q_\vtheta(\vz)} \, d\vz=\nabla_\theta\underbrace{\int q_\vtheta(\vz)\,d\vz}_{=1}=0. 
\end{align}
In conclusion, we have that
\begin{equation}
    \nabla_\vtheta\E_{q_\vtheta}(\vz)\left[\log p(\vx,\vz) - \log q_\vtheta(\vz)\right] =
    \int \left(\log p(\vx,\vz) - \log q_\vtheta(\vz)\right)\nabla_\vtheta \, q_\vtheta(\vz)d\vz.\label{eq:elbo_gradient}
\end{equation}
Sometimes, as in example \ref{box:elbo_gradient}, we can rewrite $\nabla_\vtheta q_\vtheta(\vz)$ (the gradient of the variational posterior) such that we can directly use a Monte Carlo method to estimate the integral. Later, we will cover two more general Monte Carlo-based approaches: reparameterization \citep{Kingma2013} and black-box variational inference \citep{ranganath2014black}.

\begin{featurebox}
\caption{}\label{box:elbo_gradient}
Consider a univariate Normal variational posterior parameterized by the mean $\mu$ and standard deviation $\sigma$, i.e. 
\begin{align*}
   q_\vtheta(z) &= \frac{1}{\sqrt{2\pi \sigma^2}}\exp\left(-\frac{(z-\mu)^2}{2\sigma^2}\right),\qquad \vtheta=\{\mu, \sigma\}.
\end{align*}
After some algebraic manipulations, the partial derivatives can be written as:
\begin{align*}
    \fracpartial{q_\vtheta}{\mu} &= \frac{z-\mu}{\sigma^2}\cdot q_\vtheta(z),\\
    \fracpartial{q_\vtheta}{\sigma} &= \frac{1}{\sigma}\left(\frac{(z-\mu)^2}{\sigma^2}-1\right)\cdot q_\vtheta(z).
\end{align*}
Note that $q_\vtheta(z)$ appears in both of these expressions. By inserting the above in \eqref{eq:elbo_gradient}, we thus arrive at an expectation that we can replace with a Monte Carlo estimate:
\begin{align*}
    &\nabla_\vtheta\E_{q_\vtheta}(z)\left[\log p(\vx,z) - \log q_\vtheta(z)\right] \\
    &= \int \left(\log p(\vx,z) - \log q_\vtheta(z)\right) 
    \begin{pmatrix}
    \frac{z-\mu}{\sigma^2},
    \frac{1}{\sigma}\left(\frac{(z-\mu)^2}{\sigma^2}-1\right)
    \end{pmatrix}^\top q_\vtheta(z)dz \\
    &\approx \frac{1}{L}\sum_{i=1}^L \left(\log p(\vx,z^{(i)}) - \log q_\vtheta(z^{(i)})\right)\begin{pmatrix}
    \frac{z^{(i)}-\mu}{\sigma^2},
    \frac{1}{\sigma}\left(\frac{(z^{(i)}-\mu)^2}{\sigma^2}-1\right)
    \end{pmatrix}^\top,
\end{align*}
where $z^{(i)}\sim q_\vtheta(z)$.
\end{featurebox}

\paragraph{Reparameterization}
The ``reparameterization trick'' was popularized in the work introducing the variational autoencoder \citep{Kingma2013} but the general principle has a much longer history \citep{devroye1996random}. The idea is to decouple the source of randomness from the parameters by cleverly reformulating the random variable $\vz\sim q_\vtheta(\vz)$ as a parameterized transformation $z=g_\vtheta(\rvepsilon)$ of another random variable $\rvepsilon\sim p(\rvepsilon)$ that is easy to sample. 
Effectively, this moves the randomness ``outside'' the model and makes it possible to move the gradient inside the expectation, as shown in the example below.

\begin{featurebox}
\caption{}
Suppose the variational posterior is a univariate Normal distribution parameterized by the mean $\mu$ and standard deviation $\sigma$, i.e. $q_\vtheta(z)=\mathcal{N}(z; \vtheta)$ where $\vtheta=\{\mu, \sigma\}$. This can be reparameterized as $z = g_\vtheta(\rvepsilon)= \mu + \sigma\cdot\rvepsilon$ where $\rvepsilon\sim\mathcal{N}(0, 1)$. 

Let's consider the effect this has on the expectation $\E_{q_\vtheta(z)}\left[\log z \right]$. 
\begin{enumerate}[label = (\roman*)]
    \item Original expression:
    \begin{align*}
        \E_{q_\vtheta(z)}\left[\log z\right] &= \frac{1}{\sqrt{2\pi\sigma^2}}\int \log z \exp\left(-\frac{(z-\mu)^2}{2\sigma^2}\right) dz,\\
        \nabla_\vtheta \E_{q_\vtheta(z)}\left[\log z\right] &= \int \log z \,\nabla_\vtheta q_\vtheta(z) dz\\
        &= \E_{q_\vtheta(z)}\left[\log z\cdot\begin{pmatrix}
                \left(\frac{z-\mu}{\sigma^2}\right),
                \frac{1}{\sigma}\left(\frac{(z-\mu)^2}{\sigma^2}-1\right)
            \end{pmatrix}^\top
            \right]
    \end{align*}
    where we used the expression for $\nabla_\vtheta q_\vtheta(z)$ from example \ref{box:elbo_gradient}.
    \item Reparameterized expression:
    \begin{align*}
        \E_{q_\vtheta(z)}\left[\log z\right]\bigg\vert_{z=\mu+\sigma\rvepsilon} &= \frac{1}{\sqrt{2\pi\sigma^2}}\int \log(\mu + \sigma\rvepsilon) \exp\left(-\frac{(\mu + \sigma\rvepsilon-\mu)^2}{2\sigma^2}\right) \sigma \,d\rvepsilon\\
        &= \frac{1}{\sqrt{2\pi}}\int \log(\mu + \sigma\rvepsilon) \exp\left(-\frac{\rvepsilon^2}{2}\right) \,d\rvepsilon=\E_{p(\rvepsilon)}\left[\log (\mu+\sigma\rvepsilon)\right],\\
        \nabla_\vtheta \E_{p(\rvepsilon)}\left[\log (\mu+\sigma\rvepsilon)\right]&=\E_{p(\rvepsilon)}\left[\nabla_\vtheta \log (\mu+\sigma\rvepsilon)\right]\\
        &=\E_{p(\rvepsilon)}\left[
        \begin{pmatrix}
            \frac{1}{\mu+\sigma\rvepsilon}, \frac{\rvepsilon}{\mu+\sigma\rvepsilon}
        \end{pmatrix}^\top\right].
    \end{align*}
    
\end{enumerate}

\end{featurebox}

The reparameterization trick is valid if and only if $g(\rvepsilon, \vtheta)$ is a \emph{continuous} function of $\vtheta$ for all~$\rvepsilon$ \citep{schulman2015gradient}. Further, it works in the same way as in the example above also for expectations $\E_{q_\vtheta(\vz)}\left[f(\vz)\right]$ where $f(\vz)$ is a general nonlinear function of $\vz$, 
\begin{equation}
    \nabla_\vtheta \E_{q_\vtheta(\vz)}\left[f(\vz)\right]=\nabla_\vtheta \E_{ p(\rvepsilon)}\left[f(g_\vtheta(\rvepsilon))\right]= \E_{ p(\rvepsilon)}\left[\nabla_\vtheta f(g_\vtheta(\rvepsilon))\right].
\end{equation}
By setting $f(\vz)=\log p(\vx,\vz) - \log q_\vtheta(\vz)$ we retrieve the ELBO as a special case. 


\paragraph{Amortized variational inference}
Many probabilistic models have local latent variables $\vz_i$ associated with each data point $\vx_i$. The simplest case is when the joint distribution factorizes as
\begin{equation}
    p(\vx,\vz) = \prod_{i=1}^N p(\vx_i\mid \vz_i)p(\vz_i).
\end{equation}
Suppose we use a variational posterior that factorizes accordingly,
\begin{equation}
    q_\vtheta(\vz) = \prod_{i=1}^N q_{\vtheta_i}(\vz_i),
\end{equation}
then the ELBO maximization in \eqref{eq:elbo_opt} decomposes into a sum of local ELBOs
\begin{align}
    \vtheta^* 
    &= \underset{\vtheta}{\argmin} -\sum_{i=1}^N \E_{q_{\vtheta_i}(\vz_i)}\left[  \log p(\vx_i\mid \vz_i) + \log p(\vz_i)- \log q_{\vtheta_i}(\vz_i)\right].\label{eq:local_elbo}
\end{align}
Since the optimization variables are $\vtheta=\{\vtheta_1,\ldots, \vtheta_N\}$, large datasets amount to large optimization problems, which are computationally demanding to solve. This led to the idea of \emph{amortized} variational inference \citep{rezende2014stochastic}, wherein a machine learning model (often a neural network) is trained to directly predict the solution $\vtheta^*$ of this optimization problem.

Specifically, let $\Lambda_\phi$ denote a neural network parameterized by $\phi$ that maps individual datapoints $\vx_i$ to corresponding parameters $\vtheta_i$ of the local variational posterior $q_{\vtheta_i}(\vz_i)$. This model is trained using the expression in \eqref{eq:local_elbo} as the loss function but replacing $\vtheta_i=\Lambda_\phi(\vx_i)$. Note that even though the objective function is the same, this is a form of amortized optimization \citep{amos2022tutorial} since we are now using $\phi$ as the optimization variables instead of $\vtheta$. Furthermore, the loss function is a sum over datapoints, which means that the standard machinery for training neural networks (stochastic gradient descent etc.) can be applied.
In the context of variational autoencoders, the model $\Lambda_\phi$ is referred to as the encoder, which is accompanied by a, jointly trained, decoder corresponding to the probability distribution $p(\vx\mid\vz)$ \citep{kingma2019introduction}.

\paragraph{Black-box variational inference} The reparameterization trick lets you compute the exact gradient by automatic differentiation, which is undoubtedly convenient. On the other hand, there are many models in which reparameterization is impossible. In these cases, one can instead estimate the gradient using black-box variational inference (BBVI) \citep{ranganath2014black}, which is more general yet still convenient. However, the BBVI estimator suffers from high variance.

BBVI relies on the observation that 
\begin{equation}
    \nabla_\vtheta \log q_\vtheta(\vz) = \frac{\nabla_\vtheta q_\vtheta(\vz)}{q_\vtheta(\vz)},
\end{equation}
which is sometimes referred to as the  REINFORCE trick \citep{williams1992simple}. This can be used to rewrite \eqref{eq:elbo_gradient} as
\begin{align}
    &\nabla_\vtheta\E_{q_\vtheta}(\vz)\left[\log p(\vx,\vz) - \log q_\vtheta(\vz)\right] =
    \int \left(\log p(\vx,\vz) - \log q_\vtheta(\vz)\right)\underbrace{q_\vtheta(\vz)\nabla_\vtheta \log q_\vtheta(\vz)}_{=\nabla_\vtheta q_\vtheta(\vz)}\, d\vz\nonumber\\
    &=\E_{q_\vtheta}(\vz)\left[\left(\log p(\vx,\vz) - \log q_\vtheta(\vz)\right)\nabla_\vtheta \log q_\vtheta(\vz)\right]
    \approx \frac{1}{L}\sum_{i=1}^L \left(\log p(\vx,\vz^{(i)}) - \log q_\vtheta(\vz^{(i)})\right)\nabla_\vtheta \log q_\vtheta(\vz^{(i)}).\label{eq:bbvi}
\end{align}
Since we can often use automatic differentiation to evaluate the score function $\nabla_\vtheta\log q_\vtheta(\vz)$, it appears that this reformulation resolves the problem of estimating the gradient of the ELBO from samples. The catch is, however, that this estimator often has a too high variance to be useful in practice. Arguably, the key contribution of BBVI  was to adapt two variance reduction techniques---Rao-Blackwellization and control variates---to the estimator in \eqref{eq:bbvi}.
Going into detail on these variance reduction techniques would, however, take us beyond the scope of a tutorial on the basics of variational inference. We refer the interested reader to the original work by \citet{ranganath2014black}. 


\subsection{Acknowledgments}
This work was partially supported by the Wallenberg AI, Autonomous Systems and Software Program (WASP) funded by the Knut and Alice Wallenberg Foundation. This preprint was created using the LaPreprint template (\url{https://github.com/roaldarbol/lapreprint}) by Mikkel Roald-Arb\o l.

\printbibliography

\if@endfloat\clearpage\processdelayedfloats\clearpage\fi






\end{document}